\documentclass[10pt,twocolumn,letterpaper]{article}

\usepackage{cvpr}              

\usepackage{graphicx}
\usepackage{times}
\usepackage{epsfig}
\usepackage{amsmath}
\usepackage{amssymb}
\usepackage{booktabs}
\usepackage{multirow}
\usepackage{verbatim}
\usepackage{appendix}
\usepackage{subcaption}
\usepackage{adjustbox}
\usepackage{pifont}
\usepackage{enumitem}
\usepackage{makecell}
\usepackage{float,wrapfig}
\usepackage{wasysym}
\usepackage{nicefrac}

\usepackage{color,xcolor}
\usepackage{epsfig}
\usepackage{graphicx}

\usepackage{adjustbox}
\usepackage{array}
\usepackage{booktabs}
\usepackage{colortbl}
\usepackage{float,wrapfig}
\usepackage{hhline}
\usepackage{multirow}
\usepackage{subcaption} %

\usepackage{amsmath,amsfonts,amsthm,amssymb}
\usepackage{bm}
\usepackage{nicefrac}
\usepackage{microtype}

\usepackage{changepage}
\usepackage{extramarks}
\usepackage{fancyhdr}
\usepackage{lastpage}
\usepackage{setspace}
\usepackage{soul}
\usepackage{xspace}
\usepackage{pifont}
\usepackage{cuted}
\usepackage{wrapfig}

\usepackage{multirow}
\usepackage{url}

\usepackage{algorithm}
\usepackage{algpseudocode}
\usepackage{enumitem}

\usepackage{wasysym}
\usepackage{comment}

\usepackage{comment}

\usepackage[accsupp]{axessibility} 

\usepackage[pagebackref,breaklinks,colorlinks]{hyperref}

\usepackage[capitalize]{cleveref}
\crefname{section}{Sec.}{Secs.}
\Crefname{section}{Section}{Sections}
\Crefname{table}{Table}{Tables}
\crefname{table}{Tab.}{Tabs.}

\newcommand\blfootnote[1]{%
\begingroup
\renewcommand\thefootnote{}{}\footnote{#1}%
\addtocounter{footnote}{-1}%
\endgroup
}

\def\cvprPaperID{2564} 
\def\confName{CVPR}
\def\confYear{2023}

\begin{document}

\title{CLIP2Scene: Towards Label-efficient 3D Scene Understanding by CLIP}

\author{Runnan Chen$^{1,2}$ \quad Youquan Liu$^{2,3}$ \quad Lingdong Kong$^{2,4}$ \quad Xinge Zhu$^{5}$ \quad Yuexin Ma$^{6}$ \quad Yikang Li$^{2}$\\ Yuenan Hou$^{2,\dagger}$ \quad Yu Qiao$^{2}$ \quad Wenping Wang$^{7,\dagger}$
\\[0.2ex]
\small{$^{1}$The University of Hong Kong} \quad 
\small{$^{2}$Shanghai AI Laboratory} \quad
\small{$^{3}$Hochschule Bremerhaven} \quad
\small{$^{4}$National University of Singapore}\\
\small{$^{5}$The Chinese University of Hong Kong} \quad
\small{$^{6}$ShanghaiTech University} \quad
\small{$^{7}$Texas A\&M University}
}

\def\algorithmname{CLIP2Scene}
\maketitle


\blfootnote{Symbol $^{\dagger}$ denotes the corresponding authors.}

\begin{abstract}
\vspace{-0.1cm}
Contrastive Language-Image Pre-training (CLIP) achieves promising results in 2D zero-shot and few-shot learning. Despite the impressive performance in 2D, applying CLIP to help the learning in 3D scene understanding has yet to be explored. In this paper, we make the first attempt to investigate how CLIP knowledge benefits 3D scene understanding. We propose CLIP2Scene, a simple yet effective framework that transfers CLIP knowledge from 2D image-text pre-trained models to a 3D point cloud network. We show that the pre-trained 3D network yields impressive performance on various downstream tasks, i.e., annotation-free and fine-tuning with labelled data for semantic segmentation. Specifically, built upon CLIP, we design a Semantic-driven Cross-modal Contrastive Learning framework that pre-trains a 3D network via semantic and spatial-temporal consistency regularization. For the former, we first leverage CLIP’s text semantics to select the positive and negative point samples and then employ the contrastive loss to train the 3D network. In terms of the latter, we force the consistency between the temporally coherent point cloud features and their corresponding image features. We conduct experiments on SemanticKITTI, nuScenes, and ScanNet. For the first time, our pre-trained network achieves annotation-free 3D semantic segmentation with 20.8\% and 25.08\% mIoU on nuScenes and ScanNet, respectively. When fine-tuned with 1\% or 100\% labelled data, our method significantly outperforms other self-supervised methods, with improvements of 8\% and 1\% mIoU, respectively. Furthermore, we demonstrate the generalizability for handling cross-domain datasets. Code is publicly available\footnote{\url{https://github.com/runnanchen/CLIP2Scene}.}.
\vspace{-0.5cm}
\end{abstract}

\section{Introduction}
\label{sec:introduction}

\begin{figure}
  \centerline{\includegraphics[width=0.48\textwidth]{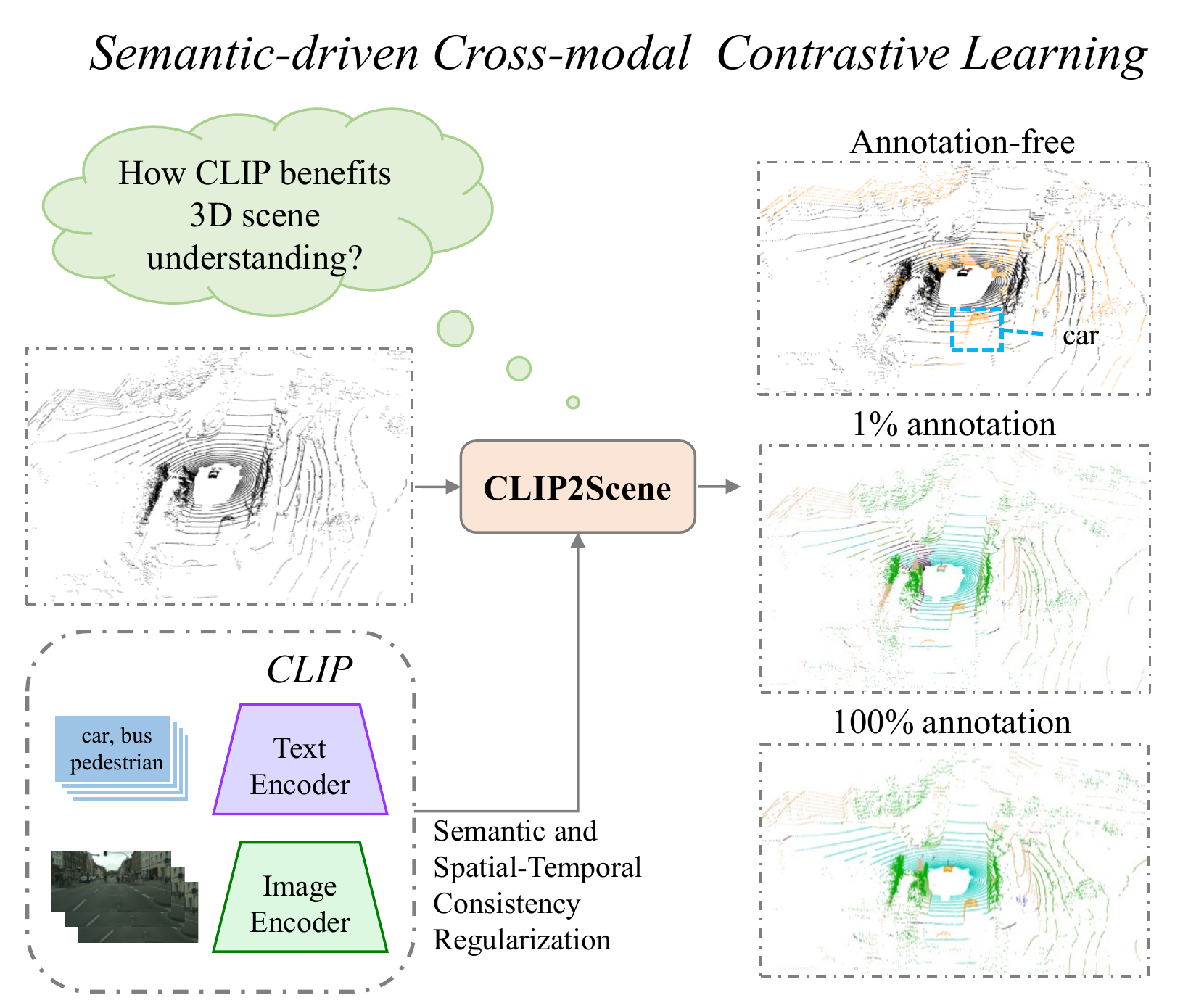}}
  \vspace{-1ex}
  \caption{We explore how  CLIP knowledge benefits 3D scene understanding. To this end, we propose CLIP2Scene, a Semantic-driven Cross-modal Contrastive Learning framework that leverages CLIP knowledge to pre-train a 3D point cloud segmentation network via semantic and spatial-temporal consistency regularization. CLIP2Scene yields impressive performance on annotation-free 3D semantic segmentation and significantly outperforms other self-supervised methods when fine-tuning on annotated data.}
  \label{fig:teaser}
  \vspace{-1ex}
\end{figure}

3D scene understanding is fundamental in autonomous driving, robot navigation, etc~\cite{3dseg_survey,3d_survey}. Current deep learning-based methods have shown inspirational performance on 3D point cloud data~\cite{qi2017pointnet,zhu2021cylindrical,pvkd2022,rpvnet,af2s3net,kong2023rethinking,hong2022dsnet}. However, some drawbacks hinder their real-world applications. The first one comes from their heavy reliance on the large collection of annotated point clouds, especially when high-quality 3D annotations are expensive to acquire ~\cite{ppkt,slidr,lasermix,kong2023conda}. Besides, they typically fail to recognize novel objects that are never seen in the training data \cite{chen2022zero,michele2021generative}. As a result, it may need extra annotation efforts to train the model on recognizing these novel objects, which is both tedious and time-consuming.

Contrastive Vision-Language Pre-training (CLIP) \cite{radford2021learning} provides a new perspective that mitigates the above issues in 2D vision. It was trained on large-scale free-available image-text pairs from websites and built vision-language correlation to achieve promising open-vocabulary recognition. MaskCLIP ~\cite{maskclip} further explores semantic segmentation based on CLIP. With minimal modifications to the CLIP pre-trained network, MaskCLIP can be directly used for the semantic segmentation of novel objects without additional training efforts. PointCLIP~\cite{PointCLIP} reveals that the zero-shot classification ability of CLIP can be generalized from the 2D image to the 3D point cloud. It perspectively projects a point cloud frame into different views of 2D depth maps that bridge the modal gap between the image and the point cloud. The above studies indicate the potential of CLIP on enhancing the 2D segmentation and 3D classification performance. However, whether and how CLIP knowledge benefits 3D scene understanding is still under-explored.

In this paper, we explore how to leverage CLIP's 2D image-text pre-learned knowledge for 3D scene understanding. Previous cross-modal knowledge distillation methods \cite{slidr,ppkt} suffer from the optimization-conflict issue, \textit{i.e.}, some of the positive pairs are regarded as negative samples for contrastive learning, leading to unsatisfactory representation learning and hammering the performance of downstream tasks. Besides, they also ignore the temporal coherence of the multi-sweep point cloud, failing to utilize the rich inter-sweep correspondence. To handle the mentioned problems, we propose a novel Semantic-driven Cross-modal Contrastive Learning framework that fully leverages CLIP's semantic and visual information to regularize a 3D network. Specifically, we propose Semantic Consistency Regularization and Spatial-Temporal Consistency Regularization. In semantic consistency regularization, we utilize CLIP's text semantics to select the positive and negative point samples for less-conflict contrastive learning. For spatial-temporal consistency regularization, we take CLIP's image pixel feature to impose a soft consistency constraint on the temporally coherent point features. Such an operation also alleviates the effects of imperfect image-to-point calibration.

We conduct several downstream tasks on the indoor and outdoor datasets to verify how the pre-trained network benefits the 3D scene understanding. The first one is annotation-free semantic segmentation. Following MaskCLIP, we place class names into multiple hand-crafted templates as prompts and average the text embeddings generated by CLIP to conduct the annotation-free segmentation. For the first time, our method achieves 20.8\% and 25.08\% mIoU annotation-free 3D semantic segmentation on the nuScenes~\cite{panoptic-nuscenes} and ScanNet~\cite{dai2017scannet} datasets without training on any labelled data. Secondly, we compare with other self-supervised methods in label-efficient learning. When fine-tuning the 3D network with 1\% or 100\% labelled data on the nuScenes dataset, our method significantly outperforms state-of-the-art self-supervised methods, with improvements of 8\% and 1\% mIoU, respectively. Besides, to verify the generalization capability, we pre-train the network on the nuScenes dataset and evaluate it on SemanticKITTI \cite{behley2019semantickitti}. Our method still significantly outperforms state-of-the-art methods.
The key contributions of our work are summarized as follows.
\begin{itemize}

\item {The first work that distils CLIP knowledge to a 3D network for 3D scene understanding.}
\item {We propose a novel Semantic-driven Cross-modal Contrastive Learning framework that pre-trains a 3D network via spatial-temporal and semantic consistency regularization.}
\item {We propose a novel Semantic-guided Spatial-Temporal Consistency Regularization that forces the consistency between the temporally coherent point cloud features and their corresponding image features.}
\item {For the first time, our method achieves promising results on annotation-free 3D scene segmentation. When fine-tuning with labelled data, our method significantly outperforms state-of-the-art self-supervised methods.}
\end{itemize}

\section{Related Work}
\label{sec:relatedwork}

\begin{figure*}
  \centerline{\includegraphics[width=1\textwidth]{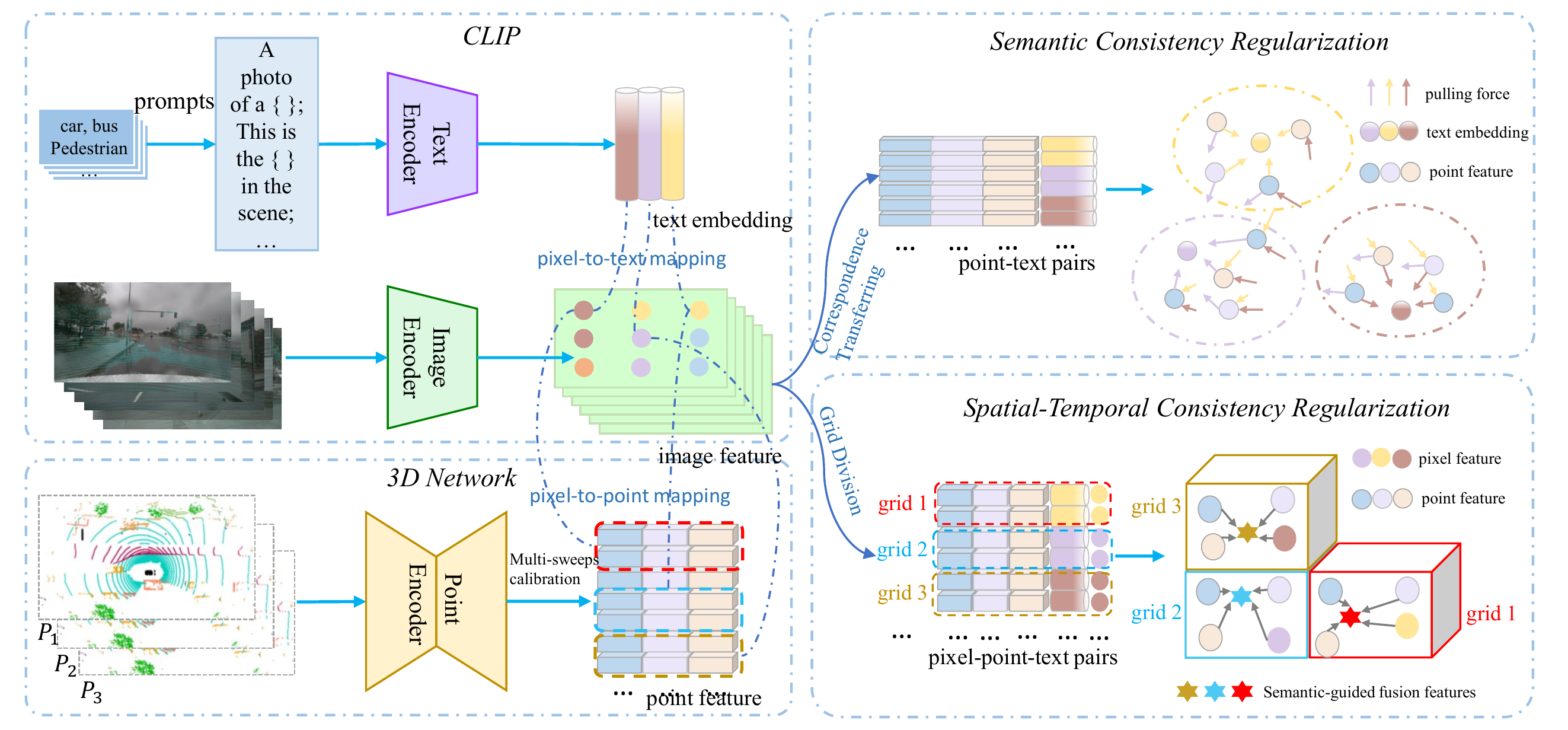}}
  \vspace{-1.8ex}
  \caption{Illustration of the Semantic-driven Cross-modal Contrastive Learning. Firstly, we obtain the text embedding $t_i$, image pixel feature $x_{i}$, and point feature $p_i$ by text encoder, image encoder, and point encoder, respectively. Secondly, we leverage CLIP knowledge to construct positive and negative samples for contrastive learning. Thus we obtain point-text pairs $\{x_i, t_i\}_{i=1}^{M}$ and all pixel-point-text pairs in a short temporal $\{\hat{x}_i^{k}, \hat{p}_i^{k}, t_i^{k}\}_{i=1,k=1}^{\hat{M},K}$. Here, $\{x_i, t_i\}_{i=1}^{M}$ and $\{\hat{x}_i^{k}, \hat{p}_i^{k}, t_i^{k}\}_{i=1,k=1}^{\hat{M},K}$ are used for Semantic Consistency Regularization and Spatial-Temporal Consistency Regularization, respectively. Lastly, we perform Semantic Consistency Regularization by pulling the point features to their corresponding text embedding and Spatial-Temporal Consistency Regularization by mimicking the temporally coherent point features to their corresponding pixel features.}
  \label{fig:framework}
  \vspace{-0.ex}
\end{figure*}

\noindent\textbf{Zero-shot Learning in 3D.} The objective of zero-shot learning (ZSL) is to recognize objects that are unseen in the training set. Many efforts have been devoted to the 2D recognition tasks \cite{Changpinyo2016SynthesizedCFzsl1, Kodirov2017SemanticAFzsl2, Zablocki2019ContextAwareZLzsl3, Mishra2018AGMzsl4, Lampert2009LearningTDzsl5, Akata2013LabelEmbeddingFAzsl6, Xian2016LatentEF_zsl7, Lampert2014AttributeBasedCF_zsl8, Bucher2017GeneratingVR_zsl9, Akata2015EvaluationOO_zsl10, Demirel2017Attributes2ClassnameAD_zsl11,Li2018DeepSS_zsl12,Gan2015ExploringSI_zsl13,xie2020pointcontrast,zhang2021self,huang2021spatio}, and few works concentrate on performing ZSL in the 3D domain \cite{cheraghian2019zero,chen2022zero,michele2021generative,cheraghian2019mitigating,cheraghian2022zero}.~\cite{cheraghian2019zero} applies ZSL to 3D tasks, where they train PointNet \cite{qi2017pointnet} on "seen" samples and test on "unseen" samples. Subsequent work~\cite{cheraghian2019mitigating} addresses the hubness problem caused by the low-quality point cloud features.~\cite{cheraghian2022zero} proposes the triplet loss to boost the performance under the transductive setting, where the "unseen" class is observed and unlabeled in the training phase.~\cite{chen2022zero} makes the first attempt to explore the transductive zero-shot segmentation for 3D scene understanding. Recently, some studies introduced CLIP into zero-shot learning. MaskCLIP~\cite{maskclip} investigates the problem of utilizing CLIP to help the 2D dense prediction tasks and exhibits encouraging zero-shot semantic segmentation performance. PointCLIP~\cite{PointCLIP} is the pioneering work that applies CLIP to 3D recognition and shows impressive performance on zero-shot and few-shot classification tasks. Our work takes a step further to investigate how the rich semantic and visual knowledge in CLIP can benefit the 3D semantic segmentation tasks.

\noindent\textbf{Self-supervised Representation Learning.} The purpose of self-supervised learning is to obtain a good representation that benefits the downstream tasks. The dominant approaches resort to contrastive learning to pre-train the network \cite{he2020momentum,grill2020bootstrap,dosovitskiy2014discriminative,doersch2015unsupervised,chen2021exploring,chen2020simple,caron2020unsupervised,chen2021referring,chen2022towards,chen2020unsupervised,chen2020unsupervised}. Recently, inspired by the success of CLIP, leveraging the pre-trained model of CLIP to the downstream tasks has raised the community's attention \cite{rao2022denseclip,yao2022detclip,rozenberszki2022language,kobayashi2022decomposing,jain2021putting}. DenseCLIP \cite{rao2022denseclip} utilizes the CLIP's pre-trained knowledge for dense image pixel prediction. DetCLIP \cite{yao2022detclip} proposes a pre-training method equipped with CLIP for open-world detection. We leverage the image-text pre-trained CLIP knowledge to help 3D scene understanding.

\noindent\textbf{Cross-modal Knowledge Distillation.} Recently, increasing studies have focused on transferring knowledge from 2D images to 3D point clouds for self-supervised representation learning~\cite {ppkt,slidr}. PPKT~\cite{ppkt} resorts to the InfoNCE loss to help the 3D network distil rich knowledge from the 2D image backbone. SLidR~\cite{slidr} further introduce the super-pixel to boost the cross-modal knowledge distillation. In this paper, we first attempt to pre-train a 3D network with CLIP's knowledge.

\section{Methodology}
\label{sec:methodology}

Considering the impressive open-vocabulary performance achieved by CLIP in image classification and segmentation, natural curiosities have been raised. Can CLIP endow the ability to a 3D network for annotation-free scene understanding? And further, will it promote the network performance when fine-tuned on labelled data? To answer the above questions, we study the cross-modal knowledge transfer of CLIP for 3D scene understanding, termed \textbf{CLIP2Scene}. Our work is a pioneer in exploiting CLIP knowledge for 3D scene understanding. In what follows, we revisit the CLIP applied in 2D open-vocabulary classification and semantic segmentation, then present our CLIP2Scene in detail. Our approach consists of three major components: Semantic Consistency Regularization, Semantic-Guided Spatial-Temporal Consistency Regularization, and Switchable Self-Training Strategy.

\begin{figure}
  \centerline{\includegraphics[width=0.5\textwidth]{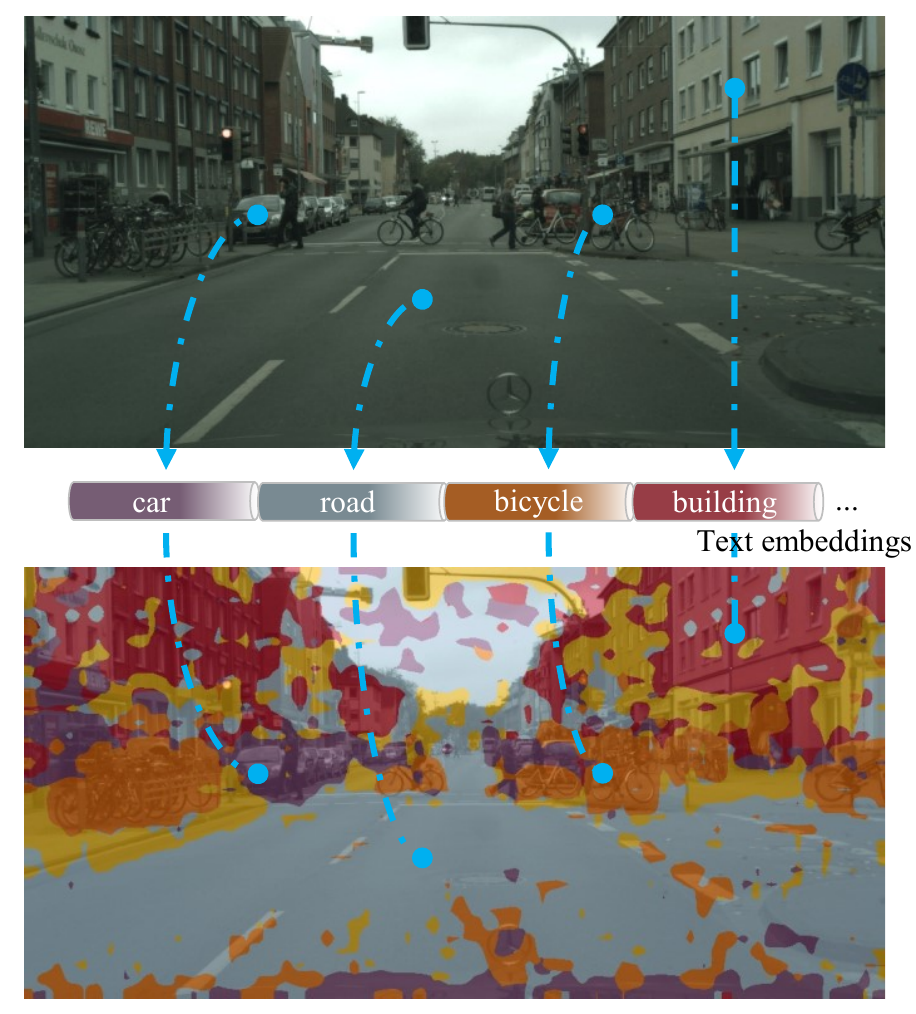}}
  \vspace{-1.5ex}
  \caption{Illustration of the image pixel-to-text mapping. The dense pixel-text correspondence $\{x_i, t_i\}_{i=1}^{M}$ is extracted by the off-the-shelf method MaskCLIP~\cite{maskclip}.}
  \label{fig:pixel_text_mapping}
  \vspace{-1.5ex}
\end{figure}

\subsection{Revisiting CLIP}

Contrastive Vision-Language Pre-training (CLIP) mitigates the following drawbacks that dominate the computer vision field: 1. Deep models need a large amount of formatted and labelled training data, which is expensive to acquire; 2. The model's generalization ability is weak, making it difficult to migrate to a new scenario with unseen objects. CLIP consists of an image encoder (ResNet \cite{he2016deep} or ViT \cite{carion2020end}) and a text encoder (Transformer \cite{vaswani2017attention}), both respectively project the image and text representation to a joint embedding space. During training, CLIP constructs positive and negative samples from 400 million image-text pairs to train both encoders with a contrastive loss, where the large-scale image-text pairs are free-available from the Internet and assumed to contain every class of images and most concepts of text. Therefore, CLIP can achieve promising open-vocabulary recognition.

For 2D zero-shot classification, CLIP first places the class name into a pre-defined template to generate the text embeddings and then encodes images to obtain image embeddings. Next, it calculates the similarities between images and text embeddings to determine the class. MaskCLIP further extends CLIP into 2D semantic segmentation. Specifically, MaskCLIP modifies the attention pooling layer of the CLIP's image encoder, thus performing pixel-level mask prediction instead of the global image-level prediction.

\subsection{CLIP2Scene}

As shown in Fig.~\ref{fig:framework}, we first leverage CLIP and 3D network to respectively extract the text embeddings, image pixel feature, and point feature. Secondly, we construct positive and negative samples based on CLIP's knowledge. Lastly, we impose Semantic Consistency Regularization by pulling the point features to their corresponding text embedding. At the same time, we apply Spatial-Temporal Consistency Regularization by forcing the consistency between temporally coherent point features and their corresponding pixel features. In what follows, we present the details and insights.

\subsubsection{Semantic Consistency Regularization}
As CLIP is pre-trained on 2D images and text, our first concern is the domain gap between 2D images and the 3D point cloud. To this end, we build dense pixel-point correspondence and transfer image knowledge to the 3D point cloud via the pixel-point pairs. Specifically, we calibrate the LiDAR point cloud with corresponding images captured by six cameras. Therefore, the dense pixel-point correspondence $\{x_i, p_i\}_{i=1}^{M}$ can be obtained accordingly, where $x_i$ and $p_i$ indicates $i$-th paired image feature and point feature, which are respectively extracted by the CLIP's image encoder and the 3D network. $M$ is the number of pairs. Note that it is an online operation and is irreverent to the image and point data augmentation. 

Previous methods \cite{slidr,ppkt} provide a promising solution to cross-modal knowledge transfer. They first construct positive pixel-point pairs $\{x_i, p_i\}_{i=1}^{M}$ and negative pairs $\{x_i, p_j\}(i \ne j)$, and then pull in the positive pairs while pushing away the negative pairs in the embedding space via the InfoNCE loss. Despite the encourageable performance of previous methods in transferring cross-modal knowledge, they are both confronted with the same optimization-conflict issue. For example, suppose $i$-th pixel $x_i$ and $j$-th point $p_j$ are in the different positions of the same instance with the same semantics. However, the InfoNCE loss will try to push them away, which is unreasonable and hammer the performance of the downstream tasks \cite{slidr}. In light of this, we propose a Semantic Consistency Regularization that leverages the CLIP's semantic information to alleviate this issue. Specifically, we generate the dense pixel-text pairs $\{x_i, t_i\}_{i=1}^{M}$ by following the off-the-shelf method MaskCLIP~\cite{maskclip} (Fig.~\ref{fig:pixel_text_mapping}), where $t_i$ is the text embedding generated from the CLIP's text encoder. Note that the pixel-text mappings are free-available from CLIP without any additional training. We then transfer pixel-text pairs to point-text pairs $\{p_i, t_i\}_{i=1}^{M}$ and utilize the text semantics to select the positive and negative point samples for contrastive learning. The objective function is as follows:

\begin{equation}\label{equ:semantic_InfoNCE}
\mathcal{L}_{S\_info} = - \sum_{c=1}^{C} \log \frac{\sum_{t_{i} \in c, p_{i}}\exp(D(t_{i}, p_{i})/\tau)}{\sum_{t_{i} \in c, t_{j} \notin c, p_{j}}\exp(D(t_{i}, p_{j})/\tau)},
\end{equation}
where $t_{i} \in c$ indicates that $t_{i}$ is generated by $c$-th classes name, and $C$ is the number of classes. $D$ denotes the scalar product operation and $\tau$ is a temperature term ($\tau > 0$).

Since the text is composed of class names placed into pre-defined templates, the text embedding represents the semantic information of the corresponding class. Therefore, those points with the same semantics will be restricted near the same text embedding, and those with different semantics will be pushed away. To this end, our Semantic Consistency Regularization causes less conflict in contrastive learning.

\subsubsection{Semantic-guided Spatial-temporal Consistency Regularization}

Besides semantic consistency regularization, we consider how image pixel features help to regularize a 3D network. The natural alternative directly pulls in the point feature with its corresponding pixel in the embedding space. However, the noise-assigned semantics of the image pixel and the imperfect pixel-point mapping hinder the downstream task's performance. To this end, we propose a novel semantic-guided Spatial-Temporal Consistency Regularization to alleviate the problem by imposing a soft constraint on points within local space and time.

\begin{figure}
  \centerline{\includegraphics[width=0.5\textwidth]{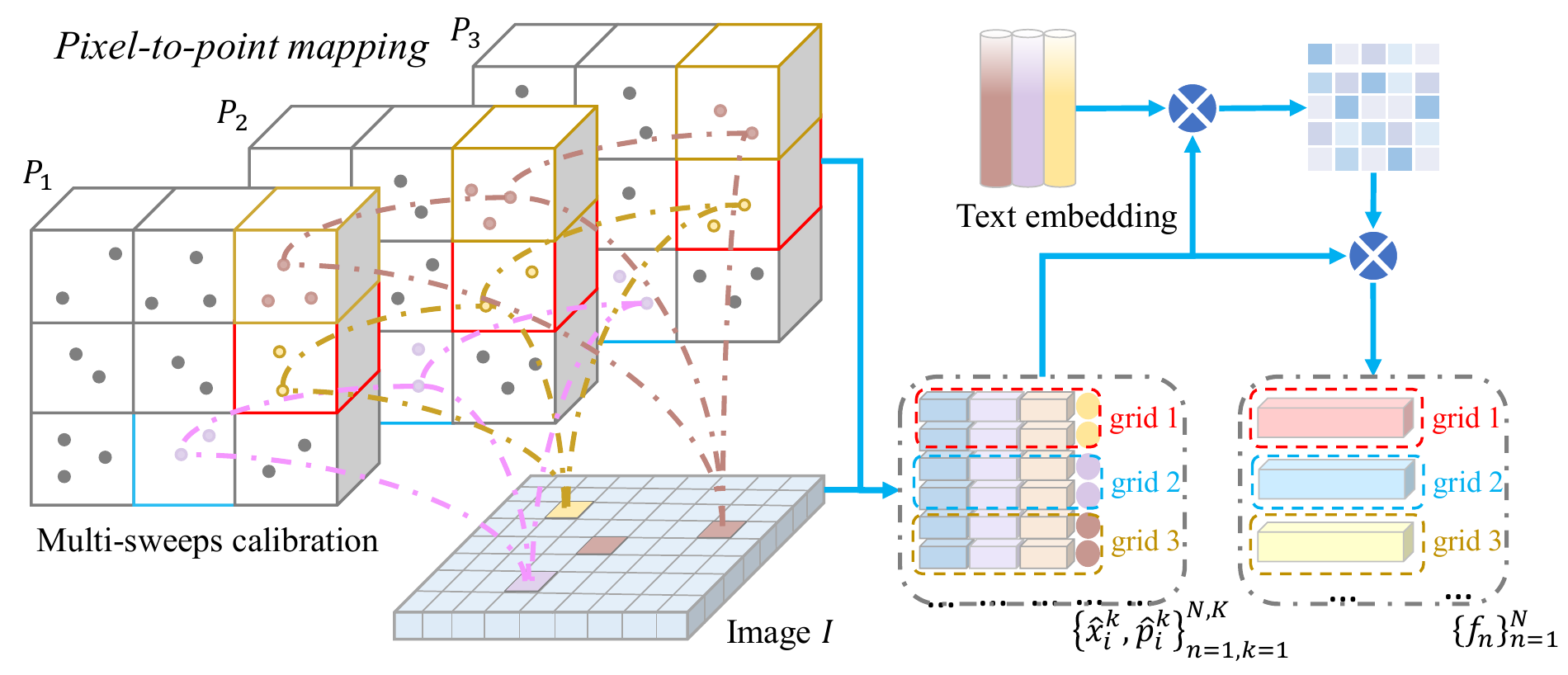}}
  \vspace{-1ex}
  \caption{Illustration of the image pixel-to-point mapping (left) and semantic-guided fusion feature generation (right). We build the grid-wise correspondence between an image $I$ and the temporally coherent LiDAR point cloud $\{P_k\}_{k=1}^{K}$ within $S$ seconds and generate semantic-guided fusion features for individual grids. Both $\{\hat{x}_i^{k}, \hat{p}_i^{k}\}_{i=1,k=1}^{\hat{M},K}$ and $\{f_{n}\}_{n=1}^{N}$ are used to perform Spatial-Temporal Consistency Regularization.}
  \label{fig:image_point_mapping}
  \vspace{-1.5ex}
\end{figure}

Specifically, given an image $I$ and temporally coherent LiDAR point cloud $\{P_k\}_{k=1}^{K}$, where $K$ is the number of sweeps within $S$ seconds. Note that the image is matched to the first frame of the point cloud $P_1$ with pixel-point pairs $\{\hat{x}_i^{1}, \hat{p}_i^{1}\}_{i=1}^{\hat{M}}$. We register the rest of the point cloud to the first frame via the calibration matrices and map them to the image (Fig.~\ref{fig:image_point_mapping}). Thus we obtain all pixel-point-text pairs in a short temporal $\{\hat{x}_i^{k}, \hat{p}_i^{k}, t_i^{k}\}_{i=1,k=1}^{\hat{M},K}$. Next, we divide the entire stitched point cloud into regular grids $\{g_n\}_{n=1}^{N}$, where the temporally coherent points are located in the same grid. We impose the spatial-temporal consistency constraint within individual grids by the following objective function:

\begin{equation}\label{equ:SSR}
\mathcal{L}_{SSR} = \sum_{g_n} \sum_{(\hat{i}, \hat{k}) \in g_n} (1 - \mathbf{sigmoid}(D(\hat{p}_{\hat{i}}^{\hat{k}}, f_n))) /N,
\end{equation}
where $(\hat{i}, \hat{k}) \in g_n$ indicates the pixel-point pair $\{\hat{x}_i^{k}, \hat{p}_i^{k}\}$ is located in the $n$-th grid. $\{f_n\}_{n=1}^{N}$ is a semantic-guided cross-modal fusion feature formulated by:
\begin{equation}\label{equ:fusion_feature}
f_n = \sum_{(\hat{i}, \hat{k}) \in g_n} a_{\hat{i}}^{\hat{k}}* \hat{x}_{\hat{i}}^{\hat{k}} + b_{\hat{i}}^{\hat{k}}* \hat{p}_{\hat{i}}^{\hat{k}},
\end{equation}
where $a_{\hat{i}}^{\hat{k}}$ and $b_{\hat{i}}^{\hat{k}}$ are attention weight calculated by:

\begin{equation}\label{equ:attention_a}
\begin{split}
a_{\hat{i}}^{\hat{k}} &= \frac{\exp(D(\hat{x}_{\hat{i}}^{\hat{k}}, t_{\hat{i}}^{1})/\lambda)}{\sum_{(\hat{i}, \hat{k}) \in g_n}\exp(D(\hat{x}_{\hat{i}}^{\hat{k}}, t_{\hat{i}}^{1})/\lambda) + \exp(D(\hat{p}_{\hat{i}}^{\hat{k}}, t_{\hat{i}}^{1})/\lambda)}, \\
b_{\hat{i}}^{\hat{k}} &= \frac{\exp(D(\hat{p}_{\hat{i}}^{\hat{k}}, t_{\hat{i}}^{1})/\lambda)}{\sum_{(\hat{i}, \hat{k}) \in g_n}\exp(D(\hat{x}_{\hat{i}}^{\hat{k}}, t_{\hat{i}}^{1})/\lambda) + \exp(D(\hat{p}_{\hat{i}}^{\hat{k}}, t_{\hat{i}}^{1})/\lambda)},
\end{split}
\end{equation}
where $\lambda$ is the temperature term.

Actually, those pixel and point features within the local grid $g_n$ are restricted near a dynamic centre $f_n$. Thus, such a soft constraint alleviates the noisy prediction and calibration error issues. At the same time, it imposes Spatio-Temporal Regularization on the temporally coherent point features.

\subsubsection{Switchable Self-training Strategy}

We combine the loss function $\mathcal{L}_{S\_info}$ and $\mathcal{L}_{SSR}$ to end-to-end train the whole network, where the CLIP's image and text encoder backbone are frozen during training. We find that method worked only when the pixel-point feature $\{x_i, p_i\}_{i=1}^{M}$ and $\{\hat{x}_i^{k}, \hat{p}_i^{k}\}_{i=1,k=1}^{\hat{M},K}$, which are used in $\mathcal{L}_{S\_info}$ and $\mathcal{L}_{SSR}$, are generated from different learnable linear layer. On top of that, we further put forward an effective strategy to promote performance. Specifically, after contrastive learning of the 3D network for a few epochs, we randomly switch the point pseudo label between the paired image pixel's pseudo label and the point's predicted label. Since different modality networks learn different feature representations, they can filter different types of error introduced by noisy pseudo labels. By this switchable operation, the error flows can be reduced by mutually \cite{han2018co}.

\section{Experiments}
\label{sec:experiments}

\paragraph{Datasets.}
We conduct extensive experiments on two large-scale outdoor LiDAR semantic segmentation datasets, \textit{i.e.}, SemanticKITTI~\cite{behley2019semantickitti} and nuScenes~\cite{caesar2020nuscenes,panoptic-nuscenes}, and one indoor dataset ScanNet \cite{dai2017scannet}. The nuScenes dataset contains 700 scenes for training, 150 scenes for validation, and 150 scenes for testing, where 16 classes are utilized for LiDAR semantic segmentation. As for SemanticKITTI, it contains 19 classes for training and evaluation. It has 22 sequences, where sequences 00 to 10, 08, and 11 to 21 are used for training, validation, and testing, respectively. ScanNet \cite{dai2017scannet}
contains 1603 scans with 20 classes, where 1201 scans are for training, 312 scans are for validation, and 100 scans are for testing.

\begin{table}[t]
  \centering
  \caption{Comparisons (mIoU) among self-supervised methods on the nuScenes~\cite{panoptic-nuscenes}, SemanticKITTI~\cite{behley2019semantickitti}, and ScanNet~\cite{dai2017scannet} \textit{val} sets.}
  \label{tab:fine_tuning}
  \vspace{-1ex}
  \scalebox{0.79}{
  \begin{tabular}{c | c c | c c | c c}
  \toprule
  \multirow{2}*{Initialization} &
  \multicolumn{2}{c}{nuScenes} & \multicolumn{2}{c}{SemanticKITTI} & \multicolumn{2}{c}{ScanNet}\\
  ~ & $1\%$ & $100\%$ & $1\%$ & $100\%$ & $5\%$ & $100\%$
  \\\midrule
  Random & $42.2$ & $69.1$ & $32.5$ & $52.1$ & $46.1$ & $63.3$
  \\\midrule
  PPKT \cite{ppkt} & $48.0$ & $70.1$ & $39.1$ & $53.1$ & $47.5$ & $64.2$
  \\
  SLidR \cite{slidr} & $48.2$ & $70.4$ & $39.6$ & $54.3$ & $47.9$ & $64.9$
  \\
  PointContrast \cite{xie2020pointcontrast}& $47.2$ & $69.2$ & $37.1$ & $52.3$ & $47.6$ & $64.5$
  \\\midrule
  CLIP2Scene & $\mathbf{56.3}$ & $\mathbf{71.5}$ & $\mathbf{42.6}$ & $\mathbf{55.0}$ & $\mathbf{48.4}$ & $\mathbf{65.1}$
  \\\bottomrule
  \end{tabular}}
\end{table}

\begin{table}[t]
  \centering
  \caption{Annotation-free 3D semantic segmentation performance (mIoU) on the nuScenes \cite{panoptic-nuscenes} and ScanNet \cite{dai2017scannet} \textit{val} sets.}
  \label{tab:scannet}
  \vspace{-1ex}
  \scalebox{0.8}{
  \begin{tabular}{c|c|c}
  \toprule
  Method & nuScenes & ScanNet
  \\
  \midrule
  CLIP2Scene & $20.80$ & $25.08$
  \\\bottomrule
  \end{tabular}}
\end{table}

\vspace{-1.5ex}
\paragraph{Implementation Details.}
We follow SLidR \cite{slidr} to pre-train the network on the nuScenes~\cite{caesar2020nuscenes,panoptic-nuscenes} dataset. The network is pre-trained on all keyframes from 600 scenes. Besides, the pre-trained network is fine-tuned on SemanticKITTI~\cite{behley2019semantickitti} to verify the generalization ability. We leverage the CLIP model to generate image features and text embedding. Following MaskCLIP, we modify the attention pooling layer of the CLIP's image encoder, thus extracting the dense pixel-text correspondences. We take SPVCNN~\cite{tang2020searching} as the 3D network to produce the point-wise feature. The framework is developed on PyTorch, where the CLIP model is frozen during training. The training time is about 40 hours for 20 epochs on two NVIDIA Tesla A100 GPUs. The optimizer is SGD with a cosine scheduler. We set the temperature $\lambda$ and $\tau$ to be 1 and 0.5, respectively. The sweep number is set to be 3 empirically. Besides, We adopt MinkowskiNet14 \cite{choy20194d} as the backbone for evaluation on the ScanNet dataset, where the number of sweeps is set to be 1 and the training epochs is 30. As for the Switchable Self-Training Strategy, we randomly switch the point supervision signal after 10 epochs. We apply several data augmentations in contrastive learning, including random rotation along the z-axis and random flip on the point cloud, random horizontal flip, and random crop-resize on the image.

\subsection{Annotation-free Semantic Segmentation}
After pre-training the network, we show the performance of the 3D network when it is not fine-tuned on any annotations (Table \ref{tab:scannet}). As no previous method reports the 3D annotation-free segmentation performance, we compare our method with different setups (Table \ref{tab:annotation_free}). In what follows, we describe the experimental settings and give insights into our method and the different settings.

\noindent\textbf{Settings.}
We conduct experiments on the nuScenes and ScanNet datasets to evaluate the annotation-free semantic segmentation performance. Following MaskCLIP \cite{maskclip}, we place the class name into 85 hand-craft prompts and feed it into the CLIP's text encoder to produce multiple text features. We then average the text features and feed the averaged features to the classifier for point-wise prediction. Besides, to explore how to effectively transfer CLIP's knowledge to the 3D network for annotation-free segmentation, We conduct the following experiments to highlight the effectiveness of different modules in our framework.

\begin{table}[t]
  \centering
  \caption{Ablation study on the nuScenes \cite{panoptic-nuscenes} \textit{val} set for annotation-free 3D semantic segmentation.}
  \vspace{-1ex}
  \label{tab:annotation_free}
  \scalebox{0.8}{
  \begin{tabular}{c|c|c}
  \toprule
  \multirow{2}*{Ablation Target} &
  \multirow{2}*{Setting} &
  \multirow{2}*{mIoU~(\%)}
  \\
  ~ & ~
  \\\midrule
  - & Baseline & $15.1$
  \\\midrule
  \multirow{4}{*}{Prompts} & nuScenes & $15.1$ \textcolor{cyan}{$_{(+0.0)}$}
  \\
  & SemanticKITTI & $13.9$ \textcolor{red}{$_{(-1.2)}$}
  \\
  & Cityscapes & $11.3$ \textcolor{red}{$_{(-3.8)}$}
  \\
  & All & $15.3$ \textcolor{cyan}{$_{(+0.2)}$}
  \\\midrule
  \multirow{3}{*}{Regularization} & {w/o} StCR & $19.8$ \textcolor{cyan}{$_{(+4.7)}$}
  \\
  & {w/o} SCR & $16.8$ \textcolor{cyan}{$_{(+1.7)}$}
  \\
  & KL & $0.0$ \textcolor{red}{$_{(-15.1)}$}
  \\\midrule
  \multirow{2}{*}{Training Strategy} & {w/o} S3 & $18.8$ \textcolor{cyan}{$_{(+3.7)}$}
  \\
  & ST & $10.1$ \textcolor{red}{$_{(-4.0)}$}
  \\\midrule
  \multirow{4}{*}{Sweeps}
  & 1 sweep & $18.7$ \textcolor{cyan}{$_{(+3.6)}$}
  \\
  & 3 sweeps & $20.8$ \textcolor{cyan}{$_{(+5.7)}$}
  \\
  & 5 sweeps & $20.6$ \textcolor{cyan}{$_{(+5.5)}$}
  \\
  & merged & $18.6$ \textcolor{cyan}{$_{(+3.5)}$}
  \\\midrule
  Full Configuration & CLIP2Scene & $\mathbf{20.8}$ \textcolor{cyan}{$_{(+5.7)}$}
  \\\bottomrule
  \end{tabular}}
\end{table}

\noindent\textbf{Baseline.} The input of the 3D semantic segmentation network is only one sweep, and we pre-train the framework via semantic consistency regularization.

\begin{figure*}
  \centerline{\includegraphics[width=1\textwidth]{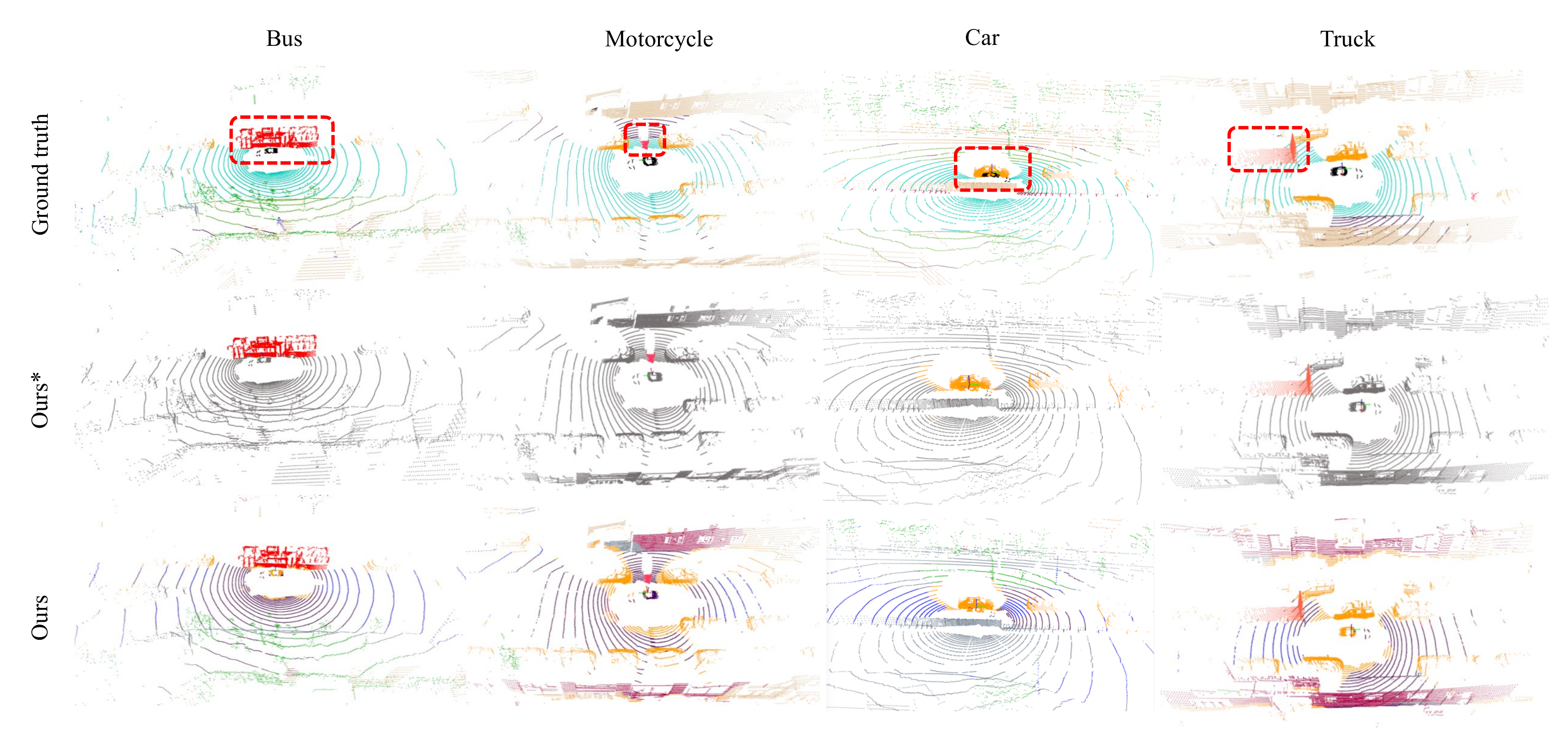}}
  \vspace{-3ex}
  \caption{Qualitative results of annotation-free semantic segmentation on nuScenes dataset. Note that we show the results of individual classes. From the left to the right column are \textit{bus}, \textit{motorcycle}, \textit{car}, and \textit{truck}, respectively. The first row [ground truth] is the annotated semantic label. The second row [ours*] is our prediction of the highlighted target. The third row [ours] is our prediction of full classes. }
  \label{fig:annotation_free}
  \vspace{-1ex}
\end{figure*}

\noindent\textbf{Prompts (nuScenes, SemanticKITTI, Cityscapes, All).} Based on the baseline, we respectively replace the nuScenes, SemanticKITTI, Cityscapes, and all class names into the prompts to produce the text embedding.

\noindent\textbf{Regularization (w/o StCR, w/o SCR, KL).} Based on the full method, we remove the Spatial-temporal Consistency Regularization (w/o StCR) and remove the Semantic Consistency Regularization (w/o SCR). Besides, we abuse both StCR and SCR and distill the image feature to the point cloud by Kullback–Leibler (KL) divergence loss.

\noindent\textbf{Training Strategies (w/o S3, ST).} We abuse the Switchable Self-Training Strategy (w/o S3) in the full method. Besides, we show the performance of only training the 3D network by their own predictions after ten epochs (ST).

\noindent\textbf{Sweeps Number (1 sweep, 3 sweeps, 5 sweeps, and merged).} We set the sweep number $K$ to be 1, 3, and 5, respectively. Besides, we also take three sweeps of the point cloud as the input to pre-train the network (merged).


\noindent\textbf{Effect of Different Prompts.}
To verify how text embedding affects the performance, we generate various text embedding by the class name from different datasets (nuScenes, SemanticKITTI, and Cityscapes) and all classes for pre-training the framework. As shown in Table \ref{tab:annotation_free}, we find that even learning with other datasets' text embedding (SemanticKITTI and Cityscapes), the 3D network could still recognize the nuScenes's objects with decent performance (13.9\% and 11.3\% mIoU, respectively). The result shows that the 3D network is capable of open-vocabulary recognition ability.

\noindent\textbf{Effect of Semantic and Spatial-temporal Consistency Regularization.}
We remove Spatial-temporal Consistency Regularization (w/o SCR) from our method. Experiments show that the performance is dramatically decreased, indicating the effectiveness of our design. Besides, we also distill the image feature to the point cloud by KL divergence loss, where the text embeddings calculate the logits. However, such a method fails to transfer the semantic information from the image. The main reason is the noise-assigned semantics of the image pixel and the imperfect pixel-point correspondence due to the calibration error.

\noindent\textbf{Effect of Switchable Self-training Strategy.}
To examine the effect of the Switchable Self-Training Strategy, we either train the network with image supervision (w/o S3) or train the 3D network by their own predictions. Both trials witness a performance drop, indicating Switchable Self-Training Strategy is efficient in cross-modal self-supervised learning.

\noindent\textbf{Effect of Sweep Numbers.}
Intuitively, the performance of our method benefits from more sweeps information. Therefore, we also show the performance when restricting sweep size to 1, 3, and 5, respectively. However, we observe that the performance of 5 sweeps is similar to 3 sweeps but is more computationally expensive. Thus, we empirically set the sweep number to be 3.

\noindent\textbf{Qualitative Evaluation.}
The qualitative evaluations of individual classes (bus, motorcycle, car, and truck) are in Fig.~\ref{fig:annotation_free}, indicating that our method is able to perceive the objects even without training on any annotated data. However, we also observe the false positive predictions around the ground truth objects. We will resolve this issue in future work.

\begin{figure*}
  \centerline{\includegraphics[width=1\textwidth]{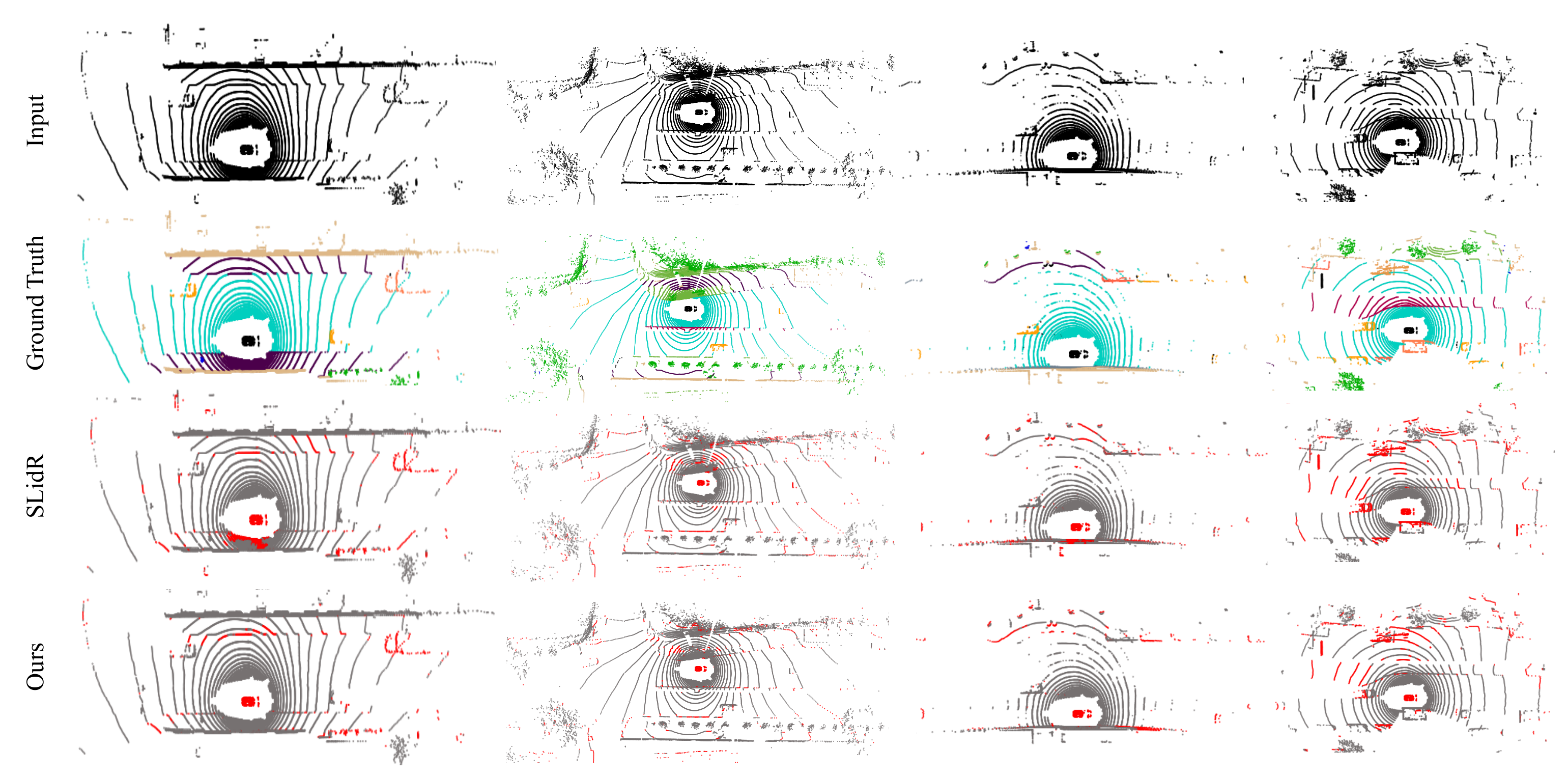}}
  \vspace{-1ex}
  \caption{Qualitative results of fine-tuning on 1\% nuScenes dataset. From the first row to the last row are the input LiDAR scan, ground truth, prediction of SLidR, and our prediction, respectively. Note that we show the results by error map, where the red point indicates the wrong prediction. Apparently, our method achieves decent performance.}
  \label{fig:visual}
\end{figure*}

\subsection{Annotation-efficient Semantic Segmentation}
The pre-trained 3D network also boosts the performance when few labeled data are available for training. We directly compare SLidR \cite{slidr}, the only published method for image-to-Lidar self-supervised representation distillation. Besides, we also compared PPKT \cite{ppkt} and PointContrast \cite{xie2020pointcontrast}. In the following, we introduce SLidR and PPKT and compare them in detail.

\noindent\textbf{PPKT.} PPKT is a cross-modal self-supervised method for the RGB-D dataset. It performs 2D-to-3D knowledge distillation via pixel-to-point contrastive loss. For a fair comparison, we use the same 3D network and training protocol but replace our semantic and Spatio-Temporal Regularization with InfoNCE loss. The framework is trained 50 epochs on 4096 randomly selected image-to-point pairs.

\noindent\textbf{SLidR.} SLidR is an image-to-Lidar self-supervised method for autonomous driving data. Compared with PPKT, it introduces image super-pixel into cross-modal self-supervised learning. For a fair comparison, we replace our loss function with their superpixel-driven contrastive loss.

\noindent\textbf{Performance.}
As shown in Table \ref{tab:fine_tuning}, our method significantly outperforms the state-of-the-art methods when fine-tuned on 1\% and 100\% nuScenes dataset, with the improvement of 8.1\% and 1.1\%, respectively. Compared with the random initialization, the improvement is 14.1\% and 2.4\%, respectively, indicating the efficiency of our Semantic-driven Cross-modal Contrastive Learning framework. The qualitative results are shown in Fig.~\ref{fig:visual}. Besides, we also verify the cross-domain generalization ability of our method. When pre-training the 3D network on the nuScenes dataset and fine-tuning on 1\% and 100\% SemanticKITTI dataset, our method significantly outperforms other state-of-the-art self-supervised methods.

\noindent\textbf{Discussions.}
PPKT and SLidR reveal that contrastive loss is promising for transferring knowledge from image to point cloud. Like self-supervised learning, constructing the positive and negative samples is vital to unsupervised cross-modal knowledge distillation. However, previous methods suffer from optimization-conflict issues, \textit{i.e.}, some negative paired samples are actually positive pairs. For example, the \textit{road} occupies a large proportion of the point cloud in a scene and is supposed to have the same semantics in the semantic segmentation task. When randomly selecting training samples, most negatively defined road-road points are actually positive. When feedforwarding such samples into contrastive learning, the contrastive loss will push them away in the embedding space, leading to unsatisfactory representation learning and hammering the downstream tasks' performance. SLidR introduces superpixel-driven contrastive learning to alleviate such issues. The motivation is that the visual representation of the image pixel and the projected points are consistent intra-superpixel. Although avoiding selecting negative image-point pairs from the same superpixel, the conflict still exists inter-superpixel. In our CLIP2Scene, we introduce the free-available dense pixel-text correspondence to alleviate the optimization conflicts. The text embedding represents the semantic information and can be used to select more reasonable training samples for contrastive learning.

Besides training sample selection, the previous method also ignores the temporal coherence of the multi-sweep point cloud. That is, for LiDAR points mapping to the same image pixel, their feature is restricted to be consistent. Besides, considering the calibration error between the LiDAR scan and the camera image. We relax the pixel-to-point mapping to image grid-to-point grid mapping for consistency regularization. To this end, our Spatial-temporal consistency regularization leads to a more rational point representation.

Last but not least, we find that randomly switching the supervision signal benefits self-supervised learning. Essentially,  different modality networks learn different feature representations. They can filter different types of errors introduced by noisy pseudo labels. By this switchable operation, the error flows can be reduced mutually.

\section{Conclusion}
\label{conclusion}
We explored how CLIP knowledge benefits 3D scene understanding in this work, termed CLIP2Scene. To efficiently transfer CLIP's image and text features to a 3D network, we propose a novel Semantic-driven Cross-modal Contrastive Learning framework including Semantic Regularization and Spatial-Temporal Regularization. For the first time, our pre-trained 3D network achieves annotation-free 3D semantic segmentation with decent performance. Besides, our method significantly outperforms state-of-the-art self-supervised methods when fine-tuning with labelled data.


{\small
\bibliographystyle{ieee_fullname}
\bibliography{egbib}
}

\end{document}